\documentclass{article} 
\usepackage{iclr2027_conference,times}

\usepackage{amsmath,amsfonts,bm}

\def\eqref#1{equation~\ref{#1}}

\def\1{\bm{1}}

\DeclareMathAlphabet{\mathsfit}{\encodingdefault}{\sfdefault}{m}{sl}
\SetMathAlphabet{\mathsfit}{bold}{\encodingdefault}{\sfdefault}{bx}{n}

\usepackage{hyperref}
\usepackage{url}
\usepackage{booktabs}       
\usepackage{amsmath}        
\usepackage{amssymb}
\usepackage{graphicx}       
\usepackage{adjustbox}
\usepackage{tabularx}
\newcolumntype{Y}{>{\raggedright\arraybackslash}X}

\title{Vault: One-Step Latent Generation with Positive-Anchored Rewards for Autonomous Driving}

\author{Yining Xing$^{1}$, Zehong Ke$^{1}$, Zhiyuan Liu$^{1}$, Jianqiang Wang$^{1}$ \\
$^{1}$Tsinghua University (tsinghua.edu.cn)
}

\iclrfinalcopy
\begin{document}

\maketitle

\lhead{Under review as a conference paper at ICLR 2027}

\begin{abstract}
End-to-end autonomous driving must balance multimodal maneuver generation against real-time inference constraints. Diffusion planners capture diverse behaviors, but their iterative denoising incurs considerable inference cost in safety-critical deployment, while one-step alternatives still imitate only the single expert demonstration, so neither family certifies both the diversity and the quality of its outputs. To address this, we propose Vault, a framework that couples one-step latent generation with sample-based reinforcement learning guidance. Built on a V-JEPA~2.1 visual representation pretrained on the full NAVSIM training imagery, Vault generates trajectory candidates in a single forward pass by drifting in a VAE latent space. During training, a reward-gated positive pool collects the maximum-scoring trajectories that the model itself discovers under the official evaluator and anchors the drift targets to these certified samples. This closes a reinforcement-learning loop purely through sample-based target construction, without policy gradients, learned reward models, or extra interaction: the generator imitates the frontier of its own certified successes, so quality and multimodality improve together under a single fixed sampling configuration. A learned scorer that predicts the official score and its sub-metrics then selects the executed trajectory, and all training-only modules are discarded at inference. On the NAVSIM benchmarks, Vault achieves state-of-the-art performance at real-time inference cost, reaching 94.6 PDMS on v1 and 91.2 EPDMS on v2.
\end{abstract}


\section{Introduction}

End-to-end (E2E) autonomous driving maps sensor inputs directly to trajectories~\citep{hu2023planning,jiang2023vad,li2024hydra}, yet urban driving is inherently multimodal~\citep{jiang2023motiondiffuser}: at a crowded unsignalized intersection, yielding, proceeding, and merging into cross-traffic are all valid, and deterministic regression, the default in most E2E planners, averages across these modes into physically implausible trajectories that straddle incompatible maneuvers. Diffusion planners capture multimodality by iterative denoising~\citep{zhong2023language,zheng2025diffusion,xing2025goalflow}, but their latency, tens to hundreds of forward passes per prediction, exceeds real-time budgets~\citep{ho2020denoising,song2021denoising}, and accelerated samplers~\citep{song2023consistency,liu2022flow,lipman2023flow} sacrifice trajectory quality for speed. A closer look at the denoising chain shows that most of it performs geometric refinement rather than the semantic shift from the unconditional prior to the scene-conditional distribution. The theory of Generative Modeling via Drifting~\citep{deng2026generative} shows that once geometric decoding is offloaded to a separate module, the semantic shift takes a single step.

One-step generation removes the latency bottleneck but leaves supervision untouched: whatever the sampler architecture, the targets remain static. Diffusion planners regress toward the single expert demonstration of each scene, so diversity must be manufactured by the noise initialization. One-step latent planners~\citep{zhang2025lap} compress the denoising chain yet inherit the same demonstration targets. Multimodality is thus borrowed from the sampling configuration rather than taught by supervision. Covering distinct maneuvers requires sweeping guidance strengths or candidate counts at inference. Moreover, the endpoint region of generation is never certified: nothing in the objective guarantees that the diverse modes are safe, drivable, or rewarded, and a static target cannot improve as the policy improves, leaving imitation quality and distributional coverage in permanent tension.

We propose Vault, which closes this gap by routing the official planning reward into the construction of the generation targets. During training, a reward-gated positive pool incrementally collects the trajectories that the model itself discovers and that the official evaluation certifies with a perfect score. The drift targets are anchored to these certified positives alongside the expert demonstration. Because every anchor is flawless by construction while the pool spans the distinct maneuvers a scene admits, the matched distribution is simultaneously high-quality and multimodal, and a single fixed sampling configuration suffices at inference. Vault realizes this design with a V-JEPA~2.1-based backbone self-pretrained on the full navtrain imagery~\citep{dauner2024navsim}, a one-step Vault Decoder, and a Vault Scorer that selects the executable candidate by predicting the official score and its sub-metrics. The pool and the VAE encoder used to construct targets are discarded at inference, leaving neither iterative sampling nor auxiliary stages in the deployed model.

Our main contributions are as follows:
\begin{itemize}
  \item \textbf{A one-step trajectory generation method.} A scene-conditioned decoder produces all trajectory proposals in a single forward pass, trained by kernel-weighted attraction to anchors and repulsion among candidates, with geometric decoding offloaded to a frozen PCA basis.
  \item \textbf{A self-evolving, sample-based reward scheme.} A reward-gated pool collects the maximum-scoring trajectories the model itself discovers under the official evaluator and anchors the drift targets to them, so supervision evolves with the policy through samples alone, without policy gradients, learned reward models, or extra interaction.
  \item \textbf{State-of-the-art NAVSIM results with fast, large-batch generation.} Under a single fixed sampling configuration, Vault surpasses all compared planners on NAVSIM v1 and v2, and its decoder emits the entire candidate set within milliseconds.
\end{itemize}

\section{Related Work}

\paragraph{E2E driving and visual representations.}
E2E driving has evolved from modular pipelines to unified architectures that plan directly from camera input~\citep{hu2023planning,jiang2023vad,jiang2026vadv2}, with recent systems introducing trajectory-vocabulary prediction, diffusion decoding, and vision-language-action modeling~\citep{li2025hydra,liao2025diffusiondrive,zhou2026autovla}. Joint-embedding pretraining has likewise been adapted to driving~\citep{wang2026drive}. Vault adopts the V-JEPA~2.1 architecture from general video modeling~\citep{assran2025v,mur2026v}, pretrains it on four-view navtrain imagery, and keeps it trainable with the planner.

\paragraph{Generative planners.}
Diffusion and flow-based planners model multiple maneuvers by iterative sampling~\citep{jiang2023motiondiffuser,zhong2023language,zheng2025diffusion,xing2025goalflow,liao2025diffusiondrive}. In general generative modeling, consistency models, rectified flow, and flow matching reduce the number of sampling steps~\citep{song2023consistency,liu2022flow,lipman2023flow}. Driving planners follow this direction: DiffusionDriveV2 truncates the denoising chain under reward constraints~\citep{zou2025diffusiondrivev2}, while LAP denoises in a single step within a trajectory-VAE space but still regresses toward the logged demonstration~\citep{zhang2025lap}. Drifting~\citep{deng2026generative} instead evolves the generated distribution across training iterations, so the deployed generator needs no denoising chain at inference. Vault instantiates this drift objective for planning and extends its anchor set with reward-filtered, self-generated trajectories.

\paragraph{Reward-guided learning and trajectory scoring.}
Self-imitation learning and reward-weighted regression improve policies by selectively reusing past successes~\citep{oh2018self}. The positive pool carries this principle into generative planning: it is self-imitation gated by a perfect official score, so the imitated successes are precisely the trajectories that the evaluator certifies. For candidate selection, Hydra-MDP, GTRS, DriveSuprim, and DrivoR learn to rank trajectory proposals with trained scorers~\citep{li2024hydra,li2025generalized,yao2026drivesuprim,kirby2026driving}, and Vault adopts a Transformer scorer similar to DrivoR without its visual register tokens. Closest to our reward-guided training, CLOVER combines evaluator-filtered pseudo-experts with scorer-mediated self-distillation~\citep{ang2026clover}. Vault differs on both axes: its drift targets are the model's own maximum-scoring trajectories rather than filtered pseudo-experts, and its scorer only ranks decoded candidates at inference rather than mediating generator training.


\section{Method}
\label{sec:method}

\subsection{Overview}
\label{sec:overview}

Vault takes surround-view images, ego state, and a navigation command, and outputs trajectory proposals from which one is executed. A ViT following the V-JEPA~2.1 architecture~\citep{assran2025v,mur2026v}, pretrained on four-view navtrain imagery and trainable with the planner, produces visual tokens $\mathbf{F}_{\text{vis}}$. Velocity and acceleration form an ego-state token $\mathbf{e}_{\text{ego}}$, and the one-hot command forms $\mathbf{e}_{\text{navi}}$. The decoder groups these state and command tokens with each noise token and attends to $\mathbf{F}_{\text{vis}}$. The scorer concatenates each candidate embedding with the ego-state and navigation tokens and attends to $\mathbf{F}_{\text{vis}}$.

\begin{figure}[t]
  \centering
  \includegraphics[width=0.78\textwidth]{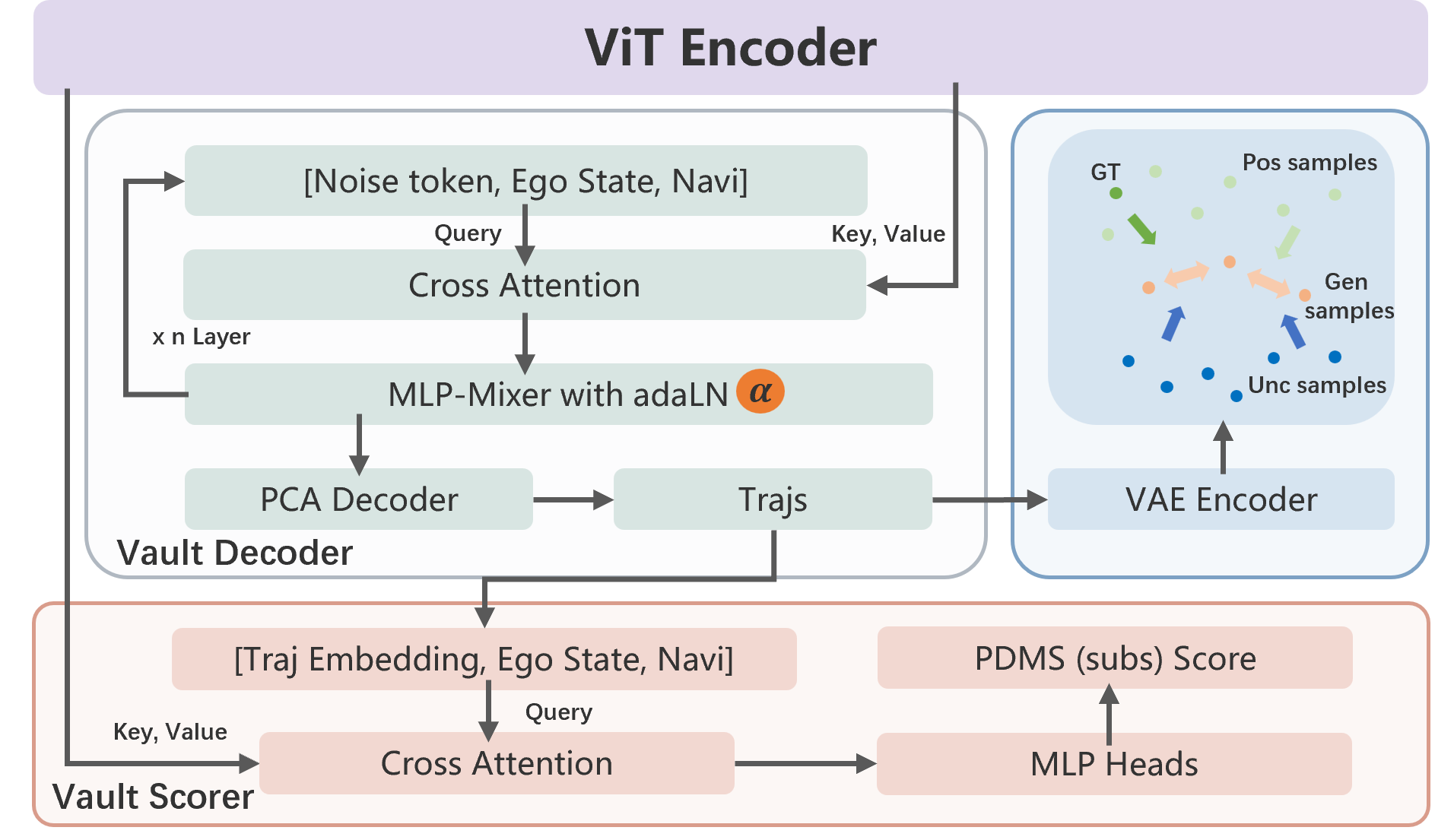}
  \caption{Overview of Vault. A V-JEPA~2.1 ViT produces visual tokens. Each noise input is grouped with ego-state and navigation tokens before visual cross-attention and PCA decoding. The scorer concatenates each trajectory embedding with the ego-state and navigation tokens and ranks candidates by predicted PDMS and sub-metrics. In the training-only inset, green denotes expert/pool anchors, blue the scene-agnostic vocabulary, and orange generated candidates. The VAE encoder and positive pool are absent at inference.}
  \label{fig:framework}
\end{figure}

The \emph{Vault Decoder} maps $N$ noise inputs to trajectory proposals in a single forward pass, through cross-attention and $\alpha$-conditioned MLP-Mixer blocks followed by a frozen PCA basis (Section~\ref{sec:decoder}). The Mixer acts within each candidate's three-token group (noise, ego state, navigation), while $N$ is a batch-like candidate axis, so the same weights accept different candidate counts. The \emph{Vault Scorer} embeds the proposals together with the ego-state and navigation tokens, attends to the visual tokens, and predicts PDMS and its sub-metrics. The top-ranked candidate is executed. Training also uses a frozen trajectory VAE encoder and a \emph{positive pool} of self-generated, maximum-scoring training trajectories (Section~\ref{sec:queue}). The expert anchor is retained independently of its score. At inference, the graph contains the visual encoder, decoder, and scorer. Figure~\ref{fig:framework} summarizes the architecture.

\subsection{One-Step Generation via Drifting}
\label{sec:decoder}

Drifting~\citep{deng2026generative} updates a generator during training rather than applying a denoising chain at inference. Vault defines this update in the latent space of a trajectory VAE and decodes output waypoints through a frozen PCA basis. A generator $f_\theta$ maps standard Gaussian noise $\boldsymbol{\epsilon}\sim\mathcal{N}(\mathbf{0},\mathbf{I})$ directly to a sample $\mathbf{x} = f_\theta(\boldsymbol{\epsilon})$. Its samples are regressed toward displaced copies along a field built from the target distribution $p$ and current generated distribution $q$,
\begin{equation}
\begin{aligned}
\mathbf{V}^{+}_{p}(\mathbf{x}) &= \frac{\mathbb{E}_{\mathbf{y}^+ \sim p}\bigl[k(\mathbf{x}, \mathbf{y}^+)\,(\mathbf{y}^+ - \mathbf{x})\bigr]}{\mathbb{E}_{\mathbf{y}^+ \sim p}\bigl[k(\mathbf{x}, \mathbf{y}^+)\bigr]}, \qquad
\mathbf{V}^{-}_{q}(\mathbf{x}) = \frac{\mathbb{E}_{\mathbf{y}^- \sim q}\bigl[k(\mathbf{x}, \mathbf{y}^-)\,(\mathbf{y}^- - \mathbf{x})\bigr]}{\mathbb{E}_{\mathbf{y}^- \sim q}\bigl[k(\mathbf{x}, \mathbf{y}^-)\bigr]},
\end{aligned}
\label{eq:field}
\end{equation}
with $\mathbf{V}_{p,q} = \mathbf{V}^{+}_{p} - \mathbf{V}^{-}_{q}$ pulling $\mathbf{x}$ toward positive samples $\mathbf{y}^+$ and away from negative samples $\mathbf{y}^-$ through the kernel $k(\mathbf{x}, \mathbf{y}) = \exp(-\|\mathbf{x} - \mathbf{y}\|_2 / \tau)$:
\begin{equation}
\mathcal{L} = \mathbb{E}_{\boldsymbol{\epsilon}}\Bigl[\bigl\| f_\theta(\boldsymbol{\epsilon}) - \mathrm{sg}\bigl(f_\theta(\boldsymbol{\epsilon}) + \mathbf{V}_{p, q_\theta}(f_\theta(\boldsymbol{\epsilon}))\bigr) \bigr\|^2\Bigr],
\label{eq:theory_loss}
\end{equation}
where $\mathrm{sg}(\cdot)$ is the stop-gradient operator. Vanishing drift is a fixed point of this update. Appendix~\ref{app:drifting} recaps the theory.

We construct the objective in the latent space of a trajectory VAE trained on reconstruction with an auxiliary maneuver-classification head, which encourages behavioral separation in the latent space (Appendix~\ref{app:drift_space} compares candidate spaces). Only the frozen VAE encoder is used during planner training, and physical trajectories come from the PCA decoder described next.

\paragraph{Vault Decoder.} Each of the $N$ noise embeddings is paired with the ego-state and navigation embeddings, giving an array of shape $B\times N\times3\times D_{\mathrm{dec}}$ for batch size $B$ and decoder feature width $D_{\mathrm{dec}}$. Each three-token group attends to visual scene features and passes through interleaved cross-attention and MLP-Mixer blocks. The Mixer operates along the within-candidate three-token and feature axes, with parameters shared across $N$, so it never mixes candidates. Other candidates influence a proposal only through the training-time repulsion field. Guidance strength $\alpha$ enters through Adaptive Layer Normalization~\citep{peebles2023scalable}, and the output is projected to waypoints $\{\boldsymbol{\tau}_i\}_{i=1}^{N}$ through a frozen PCA basis fitted on expert demonstrations.

\paragraph{Instantiated drift field.} For each scene, decoder waypoints $\boldsymbol{\tau}_i$ pass through the frozen VAE encoder to give $\mathbf{z}_i = \mathrm{VAE}_{\mathrm{enc}}(\boldsymbol{\tau}_i)$. Gradients propagate through its operations to the decoder while its weights remain fixed. The conditional anchor set $\mathcal{A} = \{\mathbf{a}_1, \dots, \mathbf{a}_M\}$ contains the expert latent $\mathbf{z}_{\text{GT}} = \mathrm{VAE}_{\mathrm{enc}}(\boldsymbol{\tau}^{\text{GT}})$ and pool positives (Section~\ref{sec:queue}). Only the latter must pass the maximum-score gate. The negatives are the other concurrently generated candidates, $\mathcal{N}_i = \{\mathbf{z}_j\}_{j \neq i}$. Euclidean latent distances, where $D$ is the VAE latent dimension, are normalized by a stopped scene scale $S_{\text{sc}} = \mathrm{sg}\bigl(\max\{\overline{d(\mathbf{z}, \mathcal{A})} / \sqrt{D},\varepsilon_s\}\bigr)$. The floor $\varepsilon_s$ keeps this scale defined when candidates and anchors coincide. The coupled assignment is
\begin{equation}
r_{ij} = \frac{e^{-d(\mathbf{z}_i,\mathbf{a}_j)/(\tau_\ell S_{\text{sc}})}}{\sum_{j'} e^{-d(\mathbf{z}_i,\mathbf{a}_{j'})/(\tau_\ell S_{\text{sc}})}}, \qquad
c_{ij} = \frac{e^{-d(\mathbf{z}_i,\mathbf{a}_j)/(\tau_\ell S_{\text{sc}})}}{\sum_{i'} e^{-d(\mathbf{z}_{i'},\mathbf{a}_j)/(\tau_\ell S_{\text{sc}})}}, \qquad
w^{+}_{ij} = \frac{\sqrt{r_{ij}\, c_{ij}}}{\sum_{j'} \sqrt{r_{ij'}\, c_{ij'}}},
\label{eq:attract}
\end{equation}
with $w^{-}_{im}$ defined identically on $\mathcal{N}_i$ after masking the self-match $m = i$. Combining the row normalization over anchors with the column normalization over candidates, each side becomes a kernel-weighted mean,
\begin{equation}
\mathbf{A}_i = \sum_{j=1}^{M} w^{+}_{ij}\, \mathbf{a}_j, \qquad
\mathbf{R}_i = \sum_{m \neq i} w^{-}_{im}\, \mathbf{z}_m .
\label{eq:repulse}
\end{equation}
The assignment permits several candidates to share an anchor while the repulsive term encourages separation. The kernel is evaluated at three bandwidths $\{\tau_\ell\} = \{1, 2, 4\}$ and the displacements are summed. Appendix~\ref{app:instantiation} relates these finite-sample, bidirectionally normalized fields to Eq.~\eqref{eq:field}.

The total field mixes a \emph{conditional} branch using the expert and pool anchors for the current scene with an \emph{unconditional} branch using a scene-agnostic vocabulary of maneuver prototypes obtained by clustering all logged expert trajectories within left-turn, straight, right-turn, and stop classes, from which $K$ prototypes are drawn each iteration (Appendix~\ref{app:vocab}):
\begin{equation}
\begin{aligned}
\mathbf{V}^{\text{cond}}_i &= \sum_\ell \bigl(\mathbf{A}_{i,\ell} - \mathbf{R}_{i,\ell}\bigr), \qquad
\mathbf{V}^{\text{unc}}_i = \sum_\ell \bigl(\bar{\mathbf{A}}_{i,\ell} - \mathbf{R}_{i,\ell}\bigr), \\
\mathbf{V}_i &= \mathbf{V}^{\text{unc}}_i + \alpha_i \bigl(\mathbf{V}^{\text{cond}}_i - \mathbf{V}^{\text{unc}}_i\bigr), \qquad
\mathcal{L}_{\text{drift}} = \frac{1}{ND} \sum_{i=1}^{N} \Bigl\| \mathbf{z}_i - \mathrm{sg}\bigl(\mathbf{z}_i + \mathbf{V}_i / \lambda_i\bigr) \Bigr\|_2^2,
\end{aligned}
\label{eq:drift_loss}
\end{equation}
where $\bar{\mathbf{A}}_{i,\ell}$ uses the scene-agnostic vocabulary instead of scene anchors and $\lambda_i = \mathrm{sg}\bigl(\max\{\sqrt{\overline{\|\mathbf{V}\|_2^2}/D},\varepsilon\}\bigr)$ normalizes the drift step, with $\varepsilon>0$ preventing division by zero. The stopped target gives a surrogate gradient proportional to $-\mathbf{V}_i/\lambda_i$ with respect to $\mathbf{z}_i$. At $\alpha=1$ the unconditional attraction cancels. Appendix~\ref{app:instantiation} analyzes the resulting fixed-point condition.

\paragraph{The $\alpha$ family.} Training samples $\alpha_i \sim \mathcal{U}[0,2]$ per candidate: $\alpha=0$ uses the scene-agnostic vocabulary, $\alpha=1$ uses scene anchors, and $\alpha>1$ extrapolates the conditional drift direction away from the unconditional one (Appendix~\ref{app:drifting}). Inference fixes $\alpha=1$ and uses one conditional forward pass.

\subsection{Reward-Gated Positive Pool}
\label{sec:queue}

\begin{figure}[t]
  \centering
  \includegraphics[width=\textwidth]{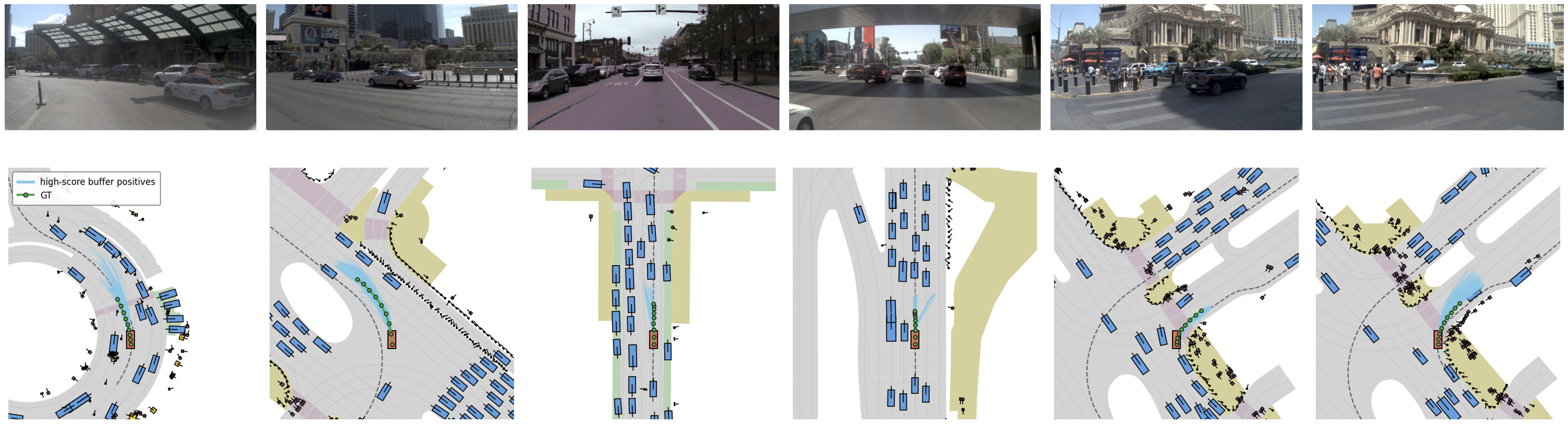}
  \caption{Reward-gated positive pool examples from six training scenes (front view above, BEV below). Green dotted marks the logged expert, and light blue marks self-generated pool trajectories admitted with the maximum evaluator score on those training scenes. Some retained branches extend farther along the route than the logged expert.}
  \label{fig:pool}
\end{figure}

During training, a pool $\mathcal{Q}$ stores latent codes of self-generated, fully feasible trajectories, each admitted by attaining the maximum score under the official evaluator on its training scene. At the end of each epoch, every newly qualifying candidate is encoded by the frozen VAE encoder and admitted to the pool, and a fixed number of stored positives is drawn per scene and iteration. This expands the supervised anchor set beyond the single logged expert. A maximum-scoring trajectory attains maximal ego progress together with maximal safety and comfort sub-scores, so the generator can imitate trajectories that advance farther than a lower-progress demonstration. Figure~\ref{fig:pool} shows examples, while the pool ablation in Section~\ref{sec:ablation_pool} assesses the system-level effect.

This is reward-filtered self-imitation: the evaluator supplies a training-only admission signal, without differentiating through the metric. Unlike a learned reward model, the gate directly uses the official score of the target benchmark, and a planner trained for a given benchmark maintains its own pool.

\subsection{Vault Scorer}
\label{sec:scorer}

Vault follows the generate-and-score paradigm~\citep{li2024hydra,yao2026drivesuprim}, adopting the Transformer-based scorer of DrivoR~\citep{kirby2026driving}. Each decoded trajectory $\boldsymbol{\tau}_i$ is embedded by a lightweight trajectory encoder, and the embedding is concatenated with the ego-state and navigation tokens $\mathbf{e}_{\text{ego}}$ and $\mathbf{e}_{\text{navi}}$. These concatenated features serve as queries in a Transformer decoder whose keys and values are the same visual tokens that the Vault Decoder attends to, and MLP heads map each output to an overall score $S_i$ and sub-scores $\{S_i^{(m)}\}$ estimating the PDMS and its sub-metrics (e.g., no at-fault collision, drivable area compliance, ego progress, time-to-collision, comfort).

Training labels come from the official scorer applied to the generated trajectories: $y_i$ is the PDMS of candidate $i$ and $y_i^{(m)}$ its sub-metrics. A pairwise hinge loss encourages score ordering, while a multi-task regression term fits the overall and component scores, $\mathcal{L}_{\text{scorer}}=\lambda_{\text{rank}}\mathcal{L}_{\text{rank}}+\lambda_{\text{mse}}\mathcal{L}_{\text{mse}}$ (Appendix~\ref{app:scorer}). At inference, the highest predicted overall score determines the executed candidate.


\section{Experiments}
\label{sec:experiments}

We report NAVSIM v1 and v2 comparisons (Section~\ref{sec:sota}), a generator swap and candidate-oracle analysis (Section~\ref{sec:drift_vs_diffusion}), a pool ablation (Section~\ref{sec:ablation_pool}), component latency (Section~\ref{sec:efficiency}), and a qualitative trajectory analysis (Section~\ref{sec:qualitative}). The generator swap matches candidate count but changes the decoder and its training objective.

\subsection{Experimental Setup}
\label{sec:training}

\paragraph{Benchmark and metrics.}
NAVSIM~\citep{dauner2024navsim} evaluates planned trajectories in non-reactive pseudo-simulation against logged scenes. We train on navtrain (1{,}192 scenarios, about 85k frames) and evaluate on navtest (about 12k scenarios), using four of eight camera views. The v1 PDM score (PDMS) combines collision avoidance (NC), drivable-area compliance (DAC), ego progress (EP), time-to-collision (TTC), and comfort (C). The v2 benchmark~\citep{Cao2025CORL} adds compliance and comfort terms to the score (EPDMS), and the v2 planner is trained separately from the v1 planner (Section~\ref{sec:training}). The original and corrected v2 evaluators have different human-reference filtering behavior, and scores computed with them should be compared within the same column (Appendix~\ref{app:metrics}).

\paragraph{Training details.}
Training has a pre-training stage followed by the main loop. Stage~0 pre-trains the ViT on navtrain, fits the trajectory VAE and PCA basis on expert demonstrations, and builds the scene-agnostic vocabulary of maneuver prototypes for the unconditional branch (Appendix~\ref{app:vocab}). The VAE encoder, PCA basis, and vocabulary are frozen thereafter, while the ViT remains trainable. The v1 and v2 planners are trained separately and share only the navtrain-pretrained ViT. Each iteration samples a scene, generates $N$ candidates with one sampled $\alpha_i\sim\mathcal{U}[0,2]$ per candidate, and assembles conditional anchors from the expert and up to $M-1$ pool positives. The planner is updated with Eq.~\eqref{eq:drift_loss} and the scorer with labels from the run's official evaluator. Every $T$ iterations, candidates are regenerated on training scenes and maximum-scoring latent codes are added to $\mathcal{Q}$. At inference, $\alpha=1$ cancels the scene-agnostic branch, and the vocabulary, pool, and VAE are not evaluated. The reported navtest scores select one of $N=1024$ proposals with the learned scorer, and at most $M=32$ scene anchors are used during training, subject to pool availability.

\subsection{Comparison with the State of the Art}
\label{sec:sota}

\begin{table*}[t]
  \caption{NAVSIM v1 navtest results (\%). Prior results are transcribed from the cited papers. The DrivoR row is the version without SimScale augmentation, as Vault also uses no simulated augmentation. Human is a reference, not a competing planner. Vault, its diffusion-decoder replacement, and the pool ablations each use 1024 proposals ranked by the learned scorer. Oracle rows select with the official evaluator and are analysis upper bounds.}
  \label{tab:navsim_v1}
  \centering
  \scriptsize
  \setlength{\tabcolsep}{6pt}
  \renewcommand{\arraystretch}{1.04}
  \begin{tabularx}{\textwidth}{Yrrrrrr}
    \toprule
    Method & NC $\uparrow$ & DAC $\uparrow$ & EP $\uparrow$ & Comfort $\uparrow$ & TTC $\uparrow$ & PDMS $\uparrow$ \\
    \midrule
    Human reference~\citep{dauner2024navsim} & 100.0 & 100.0 & 87.5 & 99.9 & 100.0 & 94.8 \\
    \midrule
    World4Drive~\citep{zheng2025world4drive} & 97.4 & 94.3 & 79.9 & 100.0 & 92.8 & 85.1 \\
    VADv2~\citep{jiang2026vadv2} & 98.1 & 94.8 & 80.6 & 100.0 & 94.3 & 86.2 \\
    DiffusionDrive~\citep{liao2025diffusiondrive} & 98.2 & 96.2 & 82.2 & 100.0 & 94.7 & 88.1 \\
    AutoVLA~\citep{zhou2026autovla} & 98.4 & 95.6 & 81.9 & 99.9 & 98.0 & 89.1 \\
    GoalFlow~\citep{xing2025goalflow} & 98.4 & 98.3 & 85.0 & 100.0 & 94.6 & 90.3 \\
    ReCogDrive~\citep{xiong2026recogdrive} & 97.9 & 97.3 & 87.3 & 100.0 & 94.9 & 90.8 \\
    Hydra-MDP++~\citep{li2025hydra} & 98.6 & 98.6 & 85.7 & 100.0 & 95.1 & 91.0 \\
    iPAD~\citep{guo2026ipad} & 98.6 & 98.3 & 88.0 & 100.0 & 94.9 & 91.7 \\
    Drive-JEPA~\citep{wang2026drive} & 99.1 & 98.2 & 91.4 & 99.9 & 96.2 & 93.7 \\
    DriveSuprim~\citep{yao2026drivesuprim} & 98.6 & 98.6 & 91.3 & 100.0 & 95.5 & 93.5 \\
    DrivoR~\citep{kirby2026driving} & 99.0 & 98.9 & 90.0 & 100.0 & 96.7 & 93.7 \\
    CLOVER~\citep{ang2026clover} & 99.1 & 99.0 & 91.7 & 100.0 & 96.9 & 94.5 \\
    \midrule
    \textbf{Vault (ours)} & 99.0 & 98.8 & 92.9 & 99.0 & 96.4 & \textbf{94.6} \\
    Vault with diffusion decoder & 99.1 & 98.6 & 86.5 & 100.0 & 96.5 & 92.0 \\
    Vault without positive pool & 98.7 & 98.8 & 91.8 & 99.9 & 95.9 & 93.9 \\
    Vault with single positive anchor & 98.3 & 98.7 & 93.0 & 99.3 & 94.7 & 93.8 \\
    \midrule
    Vault oracle best-of-64 & 99.7 & 99.8 & 98.1 & 99.9 & 98.6 & 98.4 \\
    Vault oracle best-of-256 & 99.8 & 99.8 & 98.5 & 100.0 & 99.4 & 99.0 \\
    Vault oracle best-of-1024 & 99.9 & 99.9 & 98.7 & 100.0 & 99.7 & 99.3 \\
    \bottomrule
  \end{tabularx}
\end{table*}

\paragraph{NAVSIM v1.}
With 1024 proposals and learned-scorer selection, Vault reaches 94.6 PDMS on v1 navtest (Table~\ref{tab:navsim_v1}), 0.1 above CLOVER's reported 94.5~\citep{ang2026clover} and 0.9 above the DrivoR run without SimScale augmentation~\citep{kirby2026driving}. Across five random seeds, v1 PDMS ranges from 94.52 to 94.60, all exceeding CLOVER's reported 94.5, and the reported run uses seed 1234, whose score is closest to the five-seed average. We use that DrivoR variant because Vault also has no simulated-data augmentation. Vault's EP is 92.9, above both the 91.7 of CLOVER and the 87.5 human-reference EP, while its comfort is 99.0 versus 99.9 for the human reference. Thus the selected planner makes more evaluator-measured progress than the logged human on average, with a comfort trade-off, and the human reference still has a higher aggregate PDMS of 94.8. The pool ablation below tests which part of this behavior depends on reward-gated training anchors.

\paragraph{NAVSIM v2.}
On v2 navtest, Vault obtains 88.3 EPDMS$^{*}$ under the original evaluator and 91.2 EPDMS under the corrected evaluator (Table~\ref{tab:navsim_v2}). The corrected result exceeds the paper-reported scores of SparseDriveV2 (90.1), CLOVER (90.4), and SUV (91.0)~\citep{sun2026sparsedrivev2,ang2026clover,yuan2026suv}. The 2.9-point difference between Vault's evaluator versions reflects a scoring change, not better driving. Vault has the highest mean EP among the listed planners (92.1 versus 91.1 for SparseDriveV2), above the human-reference mean of 87.4. Its lower LK (96.8), HC (97.4), and EC (76.2), however, show that higher progress does not dominate every component, and the human reference scores 94.5 EPDMS overall. The v1 and v2 planners are separately trained with their own evaluator-gated pools, so the v2 score reflects its own training metric rather than cross-metric transfer.

\begin{table*}[t]
  \caption{NAVSIM v2 navtest results (\%). EPDMS$^{*}$ uses the original evaluator and EPDMS the corrected evaluator. Sub-metrics are transcribed as reported by the cited sources, which do not always specify their evaluator version for each component. Dashes denote unavailable scores, and the human row is shown for reference.}
  \label{tab:navsim_v2}
  \centering
  \scriptsize
  \setlength{\tabcolsep}{3.8pt}
  \renewcommand{\arraystretch}{1.04}
  \begin{tabularx}{\textwidth}{Yrrrrrrrrrrr}
    \toprule
    Method & NC & DAC & DDC & TLC & EP & TTC & LK & HC & EC & EPDMS$^{*}$ & EPDMS \\
    \midrule
    Human reference~\citep{tian2026simscale,yuan2026suv} & 100.0 & 100.0 & 99.8 & 100.0 & 87.4 & 100.0 & 100.0 & 98.1 & 90.1 & 90.3 & 94.5 \\
    \midrule
    Ego Status MLP~\citep{li2024ego} & 93.1 & 77.9 & 92.7 & 99.6 & 86.0 & 91.5 & 89.4 & 98.3 & 85.4 & 64.0 & -- \\
    Transfuser~\citep{chitta2022transfuser} & 96.9 & 89.9 & 97.8 & 99.7 & 87.1 & 95.4 & 92.7 & 98.3 & 87.2 & 76.7 & -- \\
    ARTEMIS~\citep{feng2025artemis} & 98.3 & 95.1 & 98.6 & 99.8 & 81.5 & 97.4 & 96.5 & 98.3 & 98.3 & 83.1 & -- \\
    DiffusionDrive~\citep{liao2025diffusiondrive,ye2026autodrive} & 98.2 & 96.2 & 99.5 & 99.8 & 87.4 & 97.3 & 96.9 & 98.4 & 87.7 & 84.7 & 88.2 \\
    DiffusionDriveV2~\citep{zou2025diffusiondrivev2} & 97.7 & 96.6 & 99.2 & 99.8 & 88.9 & 97.2 & 96.0 & 97.8 & 91.0 & 85.5 & 87.5 \\
    Hydra-MDP++~\citep{li2025hydra} & 98.4 & 98.0 & 99.4 & 99.8 & 87.5 & 97.7 & 95.3 & 98.3 & 77.4 & 85.1 & -- \\
    DriveSuprim~\citep{yao2026drivesuprim} & 97.8 & 97.9 & 99.5 & 99.9 & 90.6 & 97.1 & 96.6 & 98.3 & 77.9 & 86.0 & -- \\
    SparseDriveV2~\citep{sun2026sparsedrivev2} & 98.1 & 98.1 & 99.6 & 99.8 & 91.1 & 97.3 & 96.9 & 98.2 & 78.4 & 86.7 & 90.1 \\
    CLOVER~\citep{ang2026clover} & 99.4 & 99.3 & 99.5 & 99.8 & 86.9 & 98.9 & 95.2 & 98.2 & 75.5 & 87.2 & 90.4 \\
    SUV~\citep{yuan2026suv} & 99.1 & 97.8 & 99.7 & 99.8 & 87.8 & 98.7 & 98.1 & 98.4 & 88.3 & -- & 91.0 \\
    \midrule
    \textbf{Vault (ours)} & 98.9 & 98.9 & 99.7 & 99.7 & 92.1 & 98.3 & 96.8 & 97.4 & 76.2 & \textbf{88.3} & \textbf{91.2} \\
    \bottomrule
  \end{tabularx}
\end{table*}

\subsection{One-Step Drift vs. Iterative Denoising}
\label{sec:drift_vs_diffusion}

We compare the one-step generator with a diffusion-decoder replacement and analyze the candidate set separately from the learned selection rule. The reported scores use the same $N=1024$ candidate budget and learned scorer. The swap changes the decoder and training objective, so it does not by itself isolate the drift objective.

\paragraph{Generator swap at matched candidate count.} With the same visual backbone, scorer, training scenes, and 1024-candidate selection budget, the diffusion decoder reaches 92.0 PDMS, versus 94.6 for Vault. Its ego progress is markedly lower, 86.5 versus 92.9, while its comfort is higher at 100.0 versus 99.0 and its safety sub-metrics are comparable, with NC 99.1 versus 99.0, DAC 98.6 versus 98.8, and TTC 96.5 versus 96.4. The diffusion replacement thus trades route progress for comfort, yielding a 2.6-point lower aggregate in this system comparison.

\paragraph{Oracle best-of-$N$.} Because the Mixer operates within each candidate rather than across candidates, the same decoder accepts different numbers of noise inputs. Separate forward passes with $N=64$, 256, and 1024 generate the three candidate sets without changing weights. Choosing the highest officially scored candidate per navtest scene yields 98.4, 99.0, and 99.3 PDMS for the full model, and 97.2, 97.8, and 98.0 for the no-pool decoder anchored only to the logged expert (Section~\ref{sec:ablation_pool}). The generated trajectories are thus already sufficiently good, reaching 99.3 PDMS under oracle selection at $N=1024$, and the full model's ceiling exceeds its no-pool counterpart at every $N$, confirming that the positive pool raises the upper bound of generation quality itself.

\subsection{Ablation: the Reward-Gated Pool}
\label{sec:ablation_pool}

The ``without positive pool'' variant keeps the backbone, scorer, and 1024-candidate selection configuration, while using only the logged expert as its conditional anchor. A pool trajectory must score PDMS $=1$ on its training scene, which entails maximal v1 EP as well as the other score components, and when the logged expert has EP $<1$, a full-score pool entry supplies a higher-EP target. Removing the pool lowers PDMS from 94.6 to 93.9. Among the v1 sub-scores, the largest positive change with the pool is EP, from 91.8 to 92.9. TTC rises from 95.9 to 96.4, NC from 98.7 to 99.0, DAC stays at 98.8, and comfort falls from 99.9 to 99.0. This pattern is consistent with more decisive, progress-oriented planning while maintaining high measured safety scores.

A complementary variant disables the pool and instead replaces the logged expert anchor with a single high-scoring stored positive. Anchored to this one sample, the planner pushes EP to 93.0, above the full model's 92.9, but its NC falls to 98.3 and its TTC to 94.7, yielding 93.8 PDMS. A single positive target thus supplies high-progress supervision yet is unstable, frequently producing safety violations, and neither it nor the logged expert alone matches the full pool. Anchoring the drift to a diverse set of certified positives appears to be what stabilizes training.

The aggregate PDMS is nonlinear, so these sub-score changes do not identify a precise causal share of its 0.7-point gain. A paired pool-versus-expert comparison and a pool-size sweep would provide a more direct test of this mechanism.

\subsection{Inference Efficiency}
\label{sec:efficiency}

On one RTX 4090, end-to-end inference at the evaluated $N=1024$ takes 91.4 ms, of which 76.3 ms is the V-JEPA encoder on sixteen $384{\times}672$ frames, 5.4 ms the decoder, and approximately 9 ms the scorer, so the encoder dominates this budget. Candidate count contributes only marginally, and reducing $N$ from 1024 to 64 shortens end-to-end inference to 82.8 ms. The reported configuration thus runs at about 11 scenes per second, meeting the 10~Hz planning requirement.

\subsection{Trajectory Generation Analysis}
\label{sec:qualitative}

\begin{figure}[t]
  \centering
  \includegraphics[width=\textwidth]{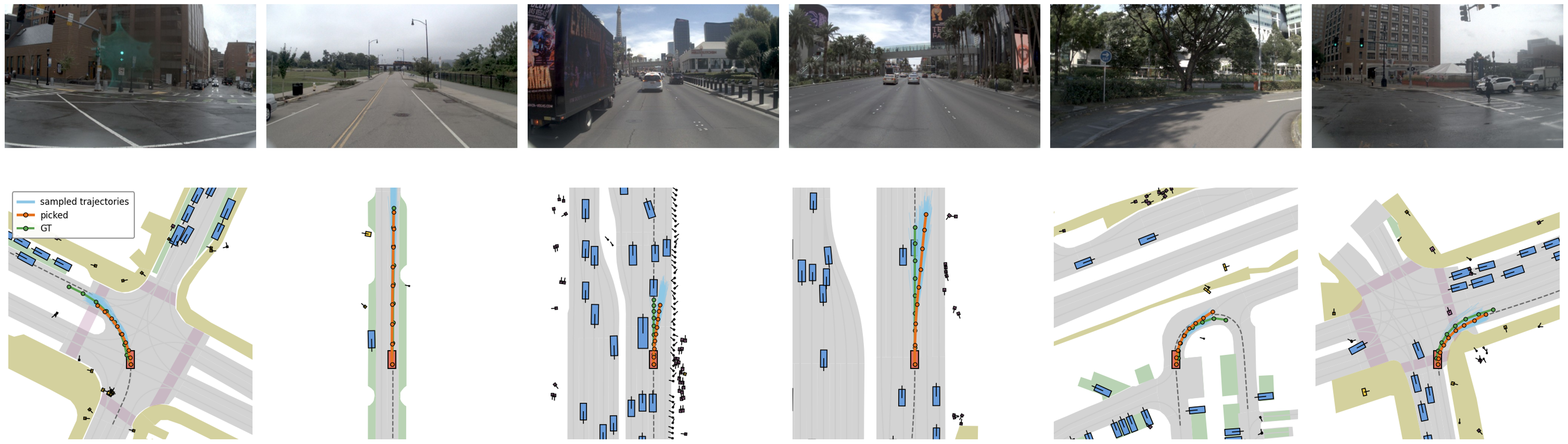}
  \caption{Generation and selection examples on navtest at $N=1024$ (front view above, BEV below). The light-blue fan marks the 1024 candidates generated in a single forward pass, orange marks the scorer selection, and green marks the logged human trajectory.}
  \label{fig:qualitative}
\end{figure}

This section presents the trajectories that the deployed model generates and selects on navtest (Figure~\ref{fig:qualitative}). We show planning results on six scenes, covering a left turn, straight driving on a narrow road, multi-lane traffic with multiple vehicles, and a right turn on a narrow road. In every scene the entire 1024-candidate set comes from a single forward pass, and its spread adapts to the scene: at decision points such as intersections and turns the fan widens into distinct maneuvers, while on constrained straight segments it contracts to a narrow band around the lane center. Diversity thus appears exactly where the scene admits multiple maneuvers, rather than being spent uniformly across all scenes.

In all of these scenes, the model generates and selects reasonable trajectories directly from the visual input. On the narrow roads, every generated trajectory stays within the road boundaries, confirming the feasibility of the generated trajectories and indicating that the aggressive full-score positives in the pool do not compromise feasibility. In the two middle multi-vehicle straight scenes, guided by the pool, the model generates right-side overtaking trajectories (the two middle panels of Figure~\ref{fig:qualitative}) and the scorer selects them, while the logged expert continues following the lead vehicle in both scenes. The source of this mode is visible in the pool itself: among the self-generated positives admitted with the maximum evaluator score (Figure~\ref{fig:pool}) are branches that overtake the lead vehicle and extend farther along the route than the logged expert. Anchoring the drift targets to these certified samples supplies the overtaking mode that the logged demonstration alone does not offer in such scenes, and the executed maneuver thereby extends beyond the expert toward a higher-progress alternative that the official evaluator certifies. In the remaining scenes the selected trajectory coincides with the logged human trajectory or deviates only by a small lateral offset through the turns, and the few candidates that leave the roadway are never selected. The self-evolving pool thus effectively guides the generation of diverse, high-quality trajectories, which underlies the EP gains observed in the pool ablation (Section~\ref{sec:ablation_pool}).

\section{Conclusion}

We presented Vault, an end-to-end planner that couples one-step trajectory generation with a self-evolving, sample-based reward scheme. The Vault Decoder produces all trajectory proposals in a single forward pass, trained by kernel-weighted attraction to anchors and repulsion among candidates in a VAE latent space, with geometric decoding offloaded to a frozen PCA basis. A reward-gated positive pool collects the maximum-scoring trajectories that the model itself discovers under the official evaluator and anchors the drift targets to them, so supervision evolves with the policy through samples alone, without policy gradients, learned reward models, or extra interaction; the generator imitates the frontier of its own certified successes, and quality and multimodality improve together under a single fixed sampling configuration. A learned scorer that predicts the official score selects the executed candidate, and all training-only modules are discarded at inference, leaving neither iterative sampling nor auxiliary stages in the deployed model. On NAVSIM, Vault reaches 94.6 PDMS on v1 and 91.2 corrected EPDMS on v2, with end-to-end inference of 91.4 ms at $N=1024$ meeting the 10~Hz planning requirement; oracle selection over the same generated sets reaches 99.3 PDMS, and the pool ablation attributes both the deployed gains and this generation ceiling to the reward-gated anchors. Testing on broader datasets beyond NAVSIM remains future work.

\subsection*{AI use statement}
The authors used a generative AI assistant for manuscript editing, consistency and formatting checks, and simulated peer review. The assistant also helped narrow the scope of several claims. They did not use generative AI for research ideation, code, experiments, or figures. The authors reviewed the AI-assisted text and take responsibility for the paper's claims, citations, numerical results, and final content.

\subsection*{Ethics statement}
This study evaluates an offline autonomous-driving planner on NAVSIM. Benchmark scores do not establish safe behavior in real traffic, and deployment would require additional validation, including rare-event and cross-domain testing.

\subsection*{Reproducibility statement}
Sections~\ref{sec:method}--\ref{sec:training} and Appendices~\ref{app:drifting}--\ref{app:additional} describe the architecture, training objective, splits, evaluator versions, reported analyses, and implementation details.


\bibliography{iclr2027_conference}
\bibliographystyle{iclr2027_conference}

\appendix

\section{Background: Generative Modeling via Drifting}
\label{app:drifting}

This section recaps the theory of Generative Modeling via Drifting~\citep{deng2026generative} to the extent needed for Section~\ref{sec:decoder}. Appendix~\ref{app:instantiation} details Vault's instantiation and its departures from the reference implementation.

\paragraph{Pushforward at training time.}
A network $f_\theta$ maps noise $\boldsymbol{\epsilon} \sim p_{\boldsymbol{\epsilon}}$ to a sample $\mathbf{x} = f_\theta(\boldsymbol{\epsilon})$. The distribution of the output is the pushforward $q_\theta = [f_\theta]_\# p_{\boldsymbol{\epsilon}}$. Diffusion and flow models evolve $q$ iteratively at inference time. Drifting instead exploits the fact that SGD training already produces a sequence of models $\{f_i\}$, and lets the training iterations themselves evolve the pushforwards $q_i = [f_i]_\# p_{\boldsymbol{\epsilon}}$, so that the deployed generator needs a single forward pass.

\paragraph{Drifting field and equilibrium.}
The evolution is governed by a field $\mathbf{V}_{p,q}$ prescribing the update $\mathbf{x}_{i+1}=\mathbf{x}_i+\mathbf{V}_{p,q_i}(\mathbf{x}_i)$ across training iterations. For an anti-symmetric field, $\mathbf{V}_{p,q}=-\mathbf{V}_{q,p}$ implies $\mathbf{V}_{p,p}=0$. The converse need not hold. \citet{deng2026generative} discuss conditions under which small drift suggests approximate matching. Vault's finite-sample field uses separate bidirectional normalizations and self-masked negatives, so it is not established to inherit the reference field's exact anti-symmetry or distribution-matching guarantees.

\paragraph{Kernel mean-shift instantiation.}
The practical field of Eq.~\eqref{eq:field} is a pair of kernel-weighted mean shifts: $\mathbf{V}^{+}_{p}$ attracts a sample toward the proximity-weighted mean of the positives $\mathbf{y}^{+} \sim p$, $\mathbf{V}^{-}_{q}$ toward that of the negatives $\mathbf{y}^{-} \sim q$, and $\mathbf{V}_{p,q} = \mathbf{V}^{+}_{p} - \mathbf{V}^{-}_{q}$ attracts by the data distribution and repels by the generated distribution. The kernel $k(\mathbf{x}, \mathbf{y}) = \exp(-\|\mathbf{x} - \mathbf{y}\|_2 / \tau)$ is realized as a softmax over samples with logits $-\|\mathbf{x} - \mathbf{y}\|_2 / \tau$, in the spirit of InfoNCE. The softmax normalizer is exactly the expectation $\mathbb{E}[k]$ in the denominator of Eq.~\eqref{eq:field}. The reference implementation additionally applies a softmax over the batch of generated samples (a bidirectional normalization that slightly improves performance in practice), and draws the negatives from the generated batch itself, masking each sample's self-match.

\paragraph{Training objective.}
The fixed-point relation $f_{\hat\theta}(\boldsymbol{\epsilon}) = f_{\hat\theta}(\boldsymbol{\epsilon}) + \mathbf{V}_{p, q_{\hat\theta}}(f_{\hat\theta}(\boldsymbol{\epsilon}))$ at equilibrium motivates the regression of Eq.~\eqref{eq:theory_loss}: the target is the current sample displaced by one drift step and frozen by the stop-gradient. The loss value equals the squared field norm, $\mathbb{E}\|\mathbf{V}\|^2$, but gradients never flow through $\mathbf{V}$ (which depends on the distribution $q_\theta$). The solver minimizes the objective indirectly by moving each sample toward its frozen, displaced self.

\paragraph{Feature-space drifting.}
The objective can be defined in any feature space: with an encoder $\phi$, the positives and negatives are $\phi(\mathbf{y}^{+})$ and $\phi(\mathbf{y}^{-})$, and the loss regresses $\phi(\mathbf{x})$ toward $\mathrm{sg}(\phi(\mathbf{x}) + \mathbf{V}(\phi(\mathbf{x})))$. The original work uses pre-trained self-supervised encoders and stresses that feature encoding is a training-time operation, absent at inference. Vault instantiates exactly this form with $\phi = \mathrm{VAE}_{\mathrm{enc}}$ (Appendix~\ref{app:drift_space} motivates this choice of space), and the kernel proximity then measures behavioral similarity between trajectories.

\paragraph{Distribution-level guidance.}
Drifting also admits a classifier-free-guidance-style behavior at training time. With class (scene) condition $c$, the negatives are drawn from a mixture $\tilde{q}(\cdot \mid c) = (1 - \gamma)\, q_\theta(\cdot \mid c) + \gamma\, p_{\text{data}}(\cdot \mid \varnothing)$ that mixes generated samples with \emph{real} unconditional data. Driving $\tilde{q}(\cdot \mid c)$ toward $p_{\text{data}}(\cdot \mid c)$ then requires the conditional generator to match
\begin{equation}
q_\theta(\cdot \mid c) = \alpha\, p_{\text{data}}(\cdot \mid c) - (\alpha - 1)\, p_{\text{data}}(\cdot \mid \varnothing), \qquad \alpha = \tfrac{1}{1 - \gamma} \ge 1,
\label{eq:extrap}
\end{equation}
so $\alpha = 1$ is pure conditional matching and $\alpha > 1$ extrapolates away from the unconditional distribution. This is a training-time mechanism. One-step inference is unaffected.

\section{Instantiating the Drift Field in Vault}
\label{app:instantiation}

This section details how Section~\ref{sec:decoder} instantiates the theory above, and collects the derivations omitted from the main text.

\subsection{Anchors, Negatives, and Scale Normalization}
\label{app:instantiation:sets}

The conditional anchor set $\mathcal{A} = \{\mathbf{z}_{\text{GT}}\} \cup \{\text{pool positives}\}$ contains one ungated expert latent and any available pool entries that pass the maximum-score gate (Section~\ref{sec:queue}). The unconditional branch instead uses the scene-agnostic vocabulary, which is not guaranteed feasible in the current scene. The generated distribution is represented by the $N$ candidates of the current pass, with the other candidates $\mathcal{N}_i = \{\mathbf{z}_j\}_{j \ne i}$ as negatives for candidate $i$.

Latent distances are uncalibrated: their absolute scale depends on the VAE and their concentration on the latent dimension $D$. Both sides of every kernel are therefore evaluated on distances divided by the scene scale
\begin{equation}
S_{\text{sc}} = \mathrm{sg}\Bigl( \max\{\overline{d(\mathbf{z}, \mathcal{A})} / \sqrt{D},\varepsilon_s\} \Bigr),
\end{equation}
where the bar averages candidate-to-anchor distances over all pairs in a scene and $\varepsilon_s>0$ is a numerical floor. This makes kernel bandwidths relative to the current latent scale. The value of the floor must be reported with the implementation. The $\sqrt{D}$ term changes how bandwidths vary with latent dimension but does not make the learned space scale-free.

\subsection{Bidirectional Assignment and Kernel Mean Shift}
\label{app:instantiation:kernel}

With anchors as an empirical measure of $p$, the row-softmax of Eq.~\eqref{eq:attract},
\begin{equation}
r_{ij} = \frac{e^{-d(\mathbf{z}_i, \mathbf{a}_j)/(\tau_\ell S_{\text{sc}})}}{\sum_{j'} e^{-d(\mathbf{z}_i, \mathbf{a}_{j'})/(\tau_\ell S_{\text{sc}})}} = \frac{k(\mathbf{z}_i, \mathbf{a}_j)}{\sum_{j'} k(\mathbf{z}_i, \mathbf{a}_{j'})},
\end{equation}
would give the empirical kernel mean $\sum_j r_{ij}\mathbf{a}_j$ and displacement $\sum_j r_{ij}\mathbf{a}_j-\mathbf{z}_i$. Vault instead uses $w^+_{ij}$ from Eq.~\eqref{eq:attract}, which also includes a column normalization. Thus $\mathbf{A}_i=\sum_j w^+_{ij}\mathbf{a}_j$ is a modified weighted mean, not exactly the empirical field in Eq.~\eqref{eq:field}. The negative side is modified analogously.

The column-softmax $c_{ij}$ normalizes over candidates for each anchor, and $\sqrt{r_{ij}c_{ij}}$ combines the two directions before row renormalization. Unlike the reference implementation of \citet{deng2026generative}, Vault normalizes the anchor and negative blocks separately. Each resulting weight vector is nonnegative and sums to one, making $\mathbf{A}_i$ and $\mathbf{R}_i$ weighted means. This changes the reference field and its theoretical properties. Many-to-one assignment permits several candidates to share an anchor. Repulsion creates a separation pressure but does not rule out a collapsed solution.

\subsection{Multi-Bandwidth Kernel}
\label{app:instantiation:bandwidth}

Small bandwidths concentrate weights near the closest anchors, while large bandwidths average more broadly. Both remain convex combinations. Vault sums displacements over $\{\tau_\ell\}=\{1,2,4\}$, following the multi-temperature idea of \citet{deng2026generative}. This reduces dependence on a single bandwidth but does not remove sensitivity to their chosen values. Any fixed-point condition for the implemented objective applies to the summed field.

\subsection{Adaptive Step Size}
\label{app:instantiation:step}

The common denominator $\lambda_i=\mathrm{sg}\bigl(\max\{\sqrt{\overline{\|\mathbf{V}\|_2^2}/D},\varepsilon\}\bigr)$ uses the batch RMS field magnitude and a positive numerical floor. For RMS above the floor, substituting the same current field into Eq.~\eqref{eq:drift_loss} yields a scalar value of exactly one. Nevertheless, the stop-gradient target gives $\partial\mathcal{L}/\partial\mathbf{z}_i=-2\mathbf{V}_i/(ND\lambda_i)$, treating the target and denominator as fixed. The equation is therefore an update surrogate, not a conventional energy whose scalar value tracks convergence. At a zero field the floor makes the update and loss well-defined and zero.

\subsection{Equilibrium Condition}
\label{app:instantiation:equilibrium}

At $\alpha=1$ the unconditional term cancels. For one bandwidth, a zero-field configuration satisfies $\mathbf{V}_i=\mathbf{A}_i-\mathbf{R}_i=0$, or
\begin{equation}
\sum_{j} w^{+}_{ij}\, \mathbf{a}_j \;=\; \sum_{m \neq i} w^{-}_{im}\, \mathbf{z}_m \qquad \text{for every candidate } i:
\label{eq:equilibrium}
\end{equation}
This equality constrains the two weighted means, not the location of $\mathbf{z}_i$ or the number of occupied modes. With multiple bandwidths, only the sum of fields needs to vanish. Consequently, coverage and quality must be established by the empirical candidate and oracle analyses rather than inferred from the fixed-point equation.

\subsection{The $\alpha$ Family}
\label{app:instantiation:alpha}

Eq.~\eqref{eq:drift_loss} trains one network over $\alpha_i\sim\mathcal{U}[0,2]$. At $\alpha=0$ the field uses the scene-agnostic vocabulary, at $\alpha=1$ it uses the current scene's expert and pool anchors, and above one it extrapolates the conditional drift direction. The algebra resembles classifier-free guidance applied to field directions, but it is not the distribution-level guidance construction of Eq.~\eqref{eq:extrap}: Vault changes the positive anchor set and leaves the co-generated negatives fixed. Values below one make the deployed value $\alpha=1$ interior to the training range. The scene-agnostic vocabulary may include trajectories infeasible in the current scene, though it cancels from the inference field at $\alpha=1$.

\subsection{Scene-Agnostic Vocabulary}
\label{app:vocab}

\begin{figure}[t]
  \centering
  \includegraphics[width=0.72\textwidth]{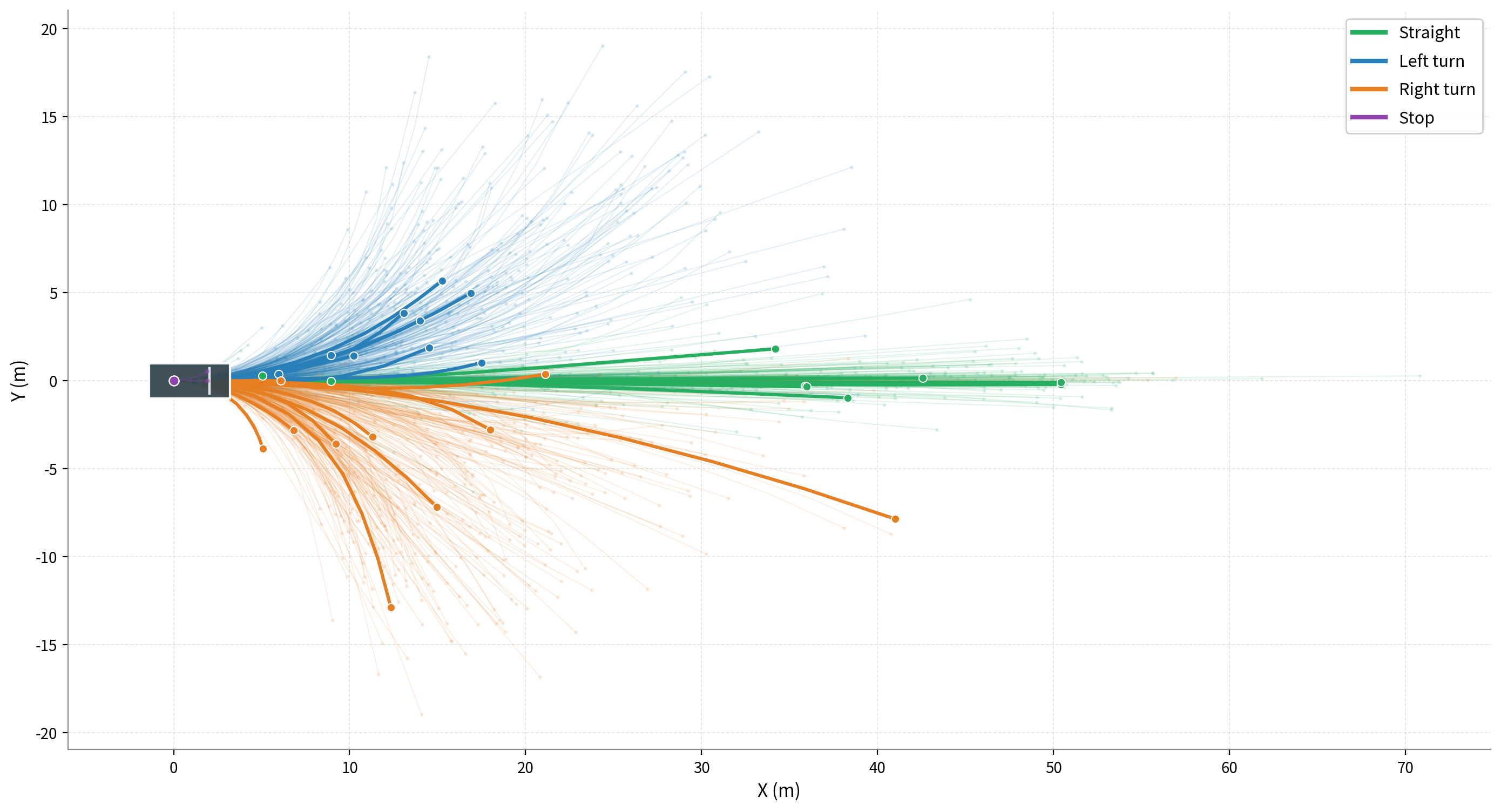}
  \caption{Scene-agnostic vocabulary in bird's-eye view, color-coded by maneuver class (left turn, straight, right turn, stop). Light trajectories are all logged expert samples, and highlighted trajectories are the per-step draws used as unconditional anchors during simulated training, allocated in a 5:5:5:1 class ratio with $K=32$.}
  \label{fig:vocab}
\end{figure}

The unconditional branch of Eq.~\eqref{eq:drift_loss} draws its anchors from a scene-agnostic vocabulary $\mathcal{V}$ built before planner training. All logged expert trajectories of navtrain are grouped by navigation command into four maneuver classes, namely left turn, straight, right turn, and stop, and each class is clustered separately, yielding 512 prototypes for each of the left-turn, straight, and right-turn classes and 32 for the stop class. At each training iteration, $K=32$ prototypes are drawn from $\mathcal{V}$ as the unconditional anchor set, allocated across the four classes in a 5:5:5:1 ratio. The vocabulary is frozen after construction, the prototypes are shared across scenes, and the branch has zero weight at the deployed $\alpha=1$. Figure~\ref{fig:vocab} illustrates the construction.

\section{Drift Space Analysis}
\label{app:drift_space}

The training-time drift is evaluated in the space of a trajectory VAE fitted with reconstruction and auxiliary maneuver-classification losses. We compare this space with physical waypoints and a PCA projection. The decoder still emits physical trajectories through its PCA basis at inference.

\begin{figure}[htbp]
    \centering
    \includegraphics[width=\textwidth]{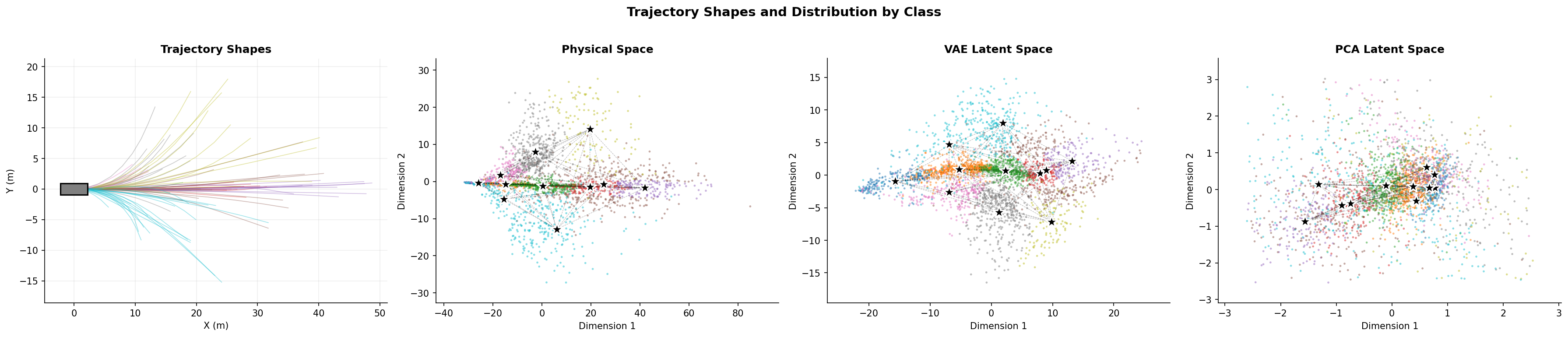}
    \caption{Trajectory shapes and 2D projections in three candidate drift spaces. The plotted sample suggests greater maneuver separation in the VAE representation, quantified by Table~\ref{tab:space_comparison}.}
    \label{fig:space_vis}
\end{figure}

For labeled maneuver classes such as left turn, right turn, and straight, Table~\ref{tab:space_comparison} reports intra-class and inter-class distances and silhouette coefficients. Distance magnitudes depend on each space's units and scaling. The silhouette coefficient is more directly comparable here.

\begin{table}[htbp]
    \centering
    \caption{Maneuver-class distance statistics for three candidate drift spaces. Distances have different units across spaces. Silhouette is dimensionless, and higher indicates better separation on these samples.}
    \label{tab:space_comparison}
    \begin{tabular*}{\textwidth}{@{\extracolsep{\fill}}lcccc@{}}
        \toprule
        \textbf{Space} & \textbf{Intra-Class} & \textbf{Inter-Class} & \textbf{Inter/Intra} & \textbf{Silhouette} \\
        \midrule
        Physical Space  & 7.98  & 28.12 & 3.52 & \phantom{-}0.025  \\
        VAE Latent Space & 4.45  & 12.99 & 2.92 & \phantom{-}0.152  \\
        PCA Space       & 2.04  & 2.18  & 1.07 & -0.458 \\
        \bottomrule
    \end{tabular*}
\end{table}

The VAE space has the highest silhouette in this analysis (0.152, versus 0.025 in physical space and $-0.458$ in PCA space), supporting its use for the drift kernel on these samples. The separation is modest and does not prove that kernel averages preserve maneuver semantics or trajectory feasibility. PCA remains the output basis, while the VAE is a training-time feature map. A further check would measure whether VAE reconstruction and PCA projection preserve rare turns and out-of-distribution expert trajectories.

\section{Vault Scorer Training Objective}
\label{app:scorer}

The scorer is trained with a pairwise hinge ranking loss on the overall score,
\begin{equation}
\mathcal{L}_{\text{rank}} = \frac{1}{\max\{1,|\mathcal{P}|\}} \sum_{(i,j) \in \mathcal{P}} \max\bigl(0,\, \delta - (S_i - S_j)\bigr),
\label{eq:rank_loss}
\end{equation}
where $\mathcal{P}=\{(i,j):y_i-y_j>\delta\}$ is the ordered set of pairs whose label difference exceeds the margin. The denominator also defines the loss for an empty pair set. With $\mathcal{M}$ denoting the sub-metrics, a regression term is
\begin{equation}
\mathcal{L}_{\text{mse}} = \frac{1}{N} \sum_{i=1}^{N} (S_i - y_i)^2 + \frac{1}{N |\mathcal{M}|} \sum_{i=1}^{N} \sum_{m \in \mathcal{M}} \bigl(S_i^{(m)} - y_i^{(m)}\bigr)^2 ,
\label{eq:mse_loss}
\end{equation}
and the total objective is $\mathcal{L}_{\text{scorer}}=\lambda_{\text{rank}}\mathcal{L}_{\text{rank}}+\lambda_{\text{mse}}\mathcal{L}_{\text{mse}}$ with $\lambda_{\text{rank}}=0.5$ and $\lambda_{\text{mse}}=1$. The pairwise term encourages correct relative ordering, while the regression term fits the overall score and auxiliary sub-metrics. Whether these terms improve top-one selection requires a separate scorer ablation.

\section{NAVSIM Metric Definitions}
\label{app:metrics}

The v1 per-scenario raw PDM score combines these components before benchmark aggregation:
\begin{equation}
s_i = \mathrm{NC}_i\,\mathrm{DAC}_i\,\frac{5\,(\mathrm{EP}_i + \mathrm{TTC}_i) + 2\,\mathrm{C}_i}{12},
\label{eq:pdms}
\end{equation}
Here NC is no at-fault collision, DAC is drivable-area compliance, EP is ego progress, TTC is time-to-collision, and C is comfort. NC and DAC are multiplicative gates, and the benchmark aggregates $s_i$ into the reported PDMS. The v2 per-scenario raw extended score adds driving-direction compliance (DDC), traffic-light compliance (TLC), lane keeping (LK), history comfort (HC), and extended comfort (EC):
\begin{equation}
q_i = \widetilde{\mathrm{NC}}_i\,\widetilde{\mathrm{DAC}}_i\,\widetilde{\mathrm{DDC}}_i\,\widetilde{\mathrm{TLC}}_i\,
\frac{5\,(\widetilde{\mathrm{EP}}_i+\widetilde{\mathrm{TTC}}_i)+2\,(\widetilde{\mathrm{LK}}_i+\widetilde{\mathrm{HC}}_i+\mathrm{EC}_i)}{16},
\label{eq:epdms}
\end{equation}
The tilde denotes the corrected official human-reference filter: for a filtered metric, a violation by the human reference suppresses the corresponding agent penalty on that scenario. It affects multiple submetrics, not just ego progress. EC is computed separately. The official evaluator performs the final split-specific aggregation. EPDMS$^{*}$ in Table~\ref{tab:navsim_v2} uses the original evaluator. EPDMS uses the corrected implementation. We compare methods only within an evaluator column.

\section{Reported Implementation Details}
\label{app:additional}

The reported planner uses four camera views over a sixteen-frame history at $384\times672$ resolution, $N=64$ candidates during training and $N=1024$ at headline inference, up to $M=32$ conditional anchors and $K=32$ unconditional vocabulary prototypes per training iteration, three kernel bandwidths $\{1,2,4\}$, and training-time $\alpha\sim\mathcal{U}[0,2]$ with inference at $\alpha=1$. Training uses AdamW with a learning rate of $10^{-4}$ for 30 epochs on eight RTX 4090 GPUs, with a global batch size of 128 realized by gradient accumulation. All reported runs use random seed 1234. The Vault Decoder has 8 layers with feature width $D_{\mathrm{dec}}=256$. The trajectory VAE has latent dimension $D=8$, and the frozen PCA output basis has 8 components. The pool is refreshed once per epoch, admitting every candidate that attains the maximum score on its training scene. The VAE encoder and PCA basis are frozen during planner training. The visual encoder remains trainable. The scorer predicts overall PDMS and component scores. End-to-end inference at $N=1024$ takes 91.4 ms on an RTX 4090, with component timings of 76.3 ms for the visual encoder, 5.4 ms for the decoder, and approximately 9 ms for the scorer.

\end{document}